\documentclass[conference]{IEEEtran}
\IEEEoverridecommandlockouts
\usepackage{cite}
\usepackage{amsmath,amssymb,amsfonts}
\usepackage{algorithmic}
\usepackage{graphicx}
\usepackage{textcomp}
\usepackage{multirow}
\usepackage{booktabs}
\usepackage{subcaption}
\usepackage[table,xcdraw]{xcolor}
\def\BibTeX{{\rm B\kern-.05em{\sc i\kern-.025em b}\kern-.08em
    T\kern-.1667em\lower.7ex\hbox{E}\kern-.125emX}}
\begin{document}
\title{AKVQ-VL: Attention-Aware KV Cache Adaptive 2-Bit Quantization for Vision-Language Models}

\author{
 Zunhai Su\textsuperscript{1},
Wang Shen\textsuperscript{2}, 
Linge Li\textsuperscript{2}, 
Zhe Chen\textsuperscript{2},
Hanyu Wei\textsuperscript{1},
Huangqi Yu\textsuperscript{2},
Kehong Yuan\textsuperscript{1}\textsuperscript{†}\thanks{\textsuperscript{†} Corresponding author.},
\\
 \textsuperscript{1}Tsinghua University, \textsuperscript{2}Huawei Technologies Co., Ltd
\\
\{zh-su23,wei-hy23\}@mails.tsinghua.edu.cn \\
\{shenwang1,lilinge,chenzhe49,yuhuangqi\}@huawei.com,
yuankh@sz.tsinghua.edu.cn
}

\maketitle
\begin{abstract}
Vision-language models (VLMs) show remarkable performance in multimodal tasks.
However, excessively long multimodal inputs lead to oversized Key-Value (KV) caches, resulting in significant memory consumption and I/O bottlenecks.
Previous KV quantization methods for Large Language Models (LLMs) may alleviate these issues but overlook the attention saliency differences of multimodal tokens, resulting in suboptimal performance.
In this paper, we investigate the attention-aware token saliency patterns in VLM and propose AKVQ-VL.
AKVQ-VL leverages the proposed Text-Salient Attention (TSA) and Pivot-Token-Salient Attention (PSA) patterns to adaptively allocate bit budgets.
Moreover, achieving extremely low-bit quantization requires effectively addressing outliers in KV tensors.
AKVQ-VL utilizes the Walsh-Hadamard transform (WHT) to construct outlier-free KV caches, thereby reducing quantization difficulty.
Evaluations of 2-bit quantization on 12 long-context and multimodal tasks demonstrate that AKVQ-VL maintains or even improves accuracy, outperforming LLM-oriented methods.
AKVQ-VL can reduce peak memory usage by 2.13×, support up to 3.25× larger batch sizes and 2.46× throughput.
\end{abstract}
\begin{IEEEkeywords}
Vision-Language Models, Key-Value Cache, Low-Bit Quantization, Attention-Aware
\end{IEEEkeywords}

\section{Introduction}
The rapid growth of multimedia content has made the processing of diverse data modalities a key focus in AI research \cite{bayoudh2022survey}.
Vision-language models (VLMs), built upon large language models (LLMs), harness the advanced capabilities of LLM to tackle a wide range of vision-related multimodal tasks \cite{zhang2024vision}.
During LLM inference, the Key-Value (KV) cache mechanism improves efficiency by storing KV pairs computed by the self-attention layer in each Transformer block \cite{vaswani2017attention}, thereby avoiding redundant computations.
Although capable of processing visual representations, VLM face the challenge of managing excessively long and redundant sequences generated by multiple images, high-resolution visuals, or multi-frame videos \cite{he2024zipvl,wan2024look}. 
An oversized KV cache leads to significant memory consumption and I/O bottlenecks.
Previous studies\cite{liu2024kivi, duanmu2024skvq,hooper2024kvquant} have explored low-bit quantization techniques to compress the KV cache in LLM.
\begin{figure}[t]
    \centering
\includegraphics[width=1\linewidth]{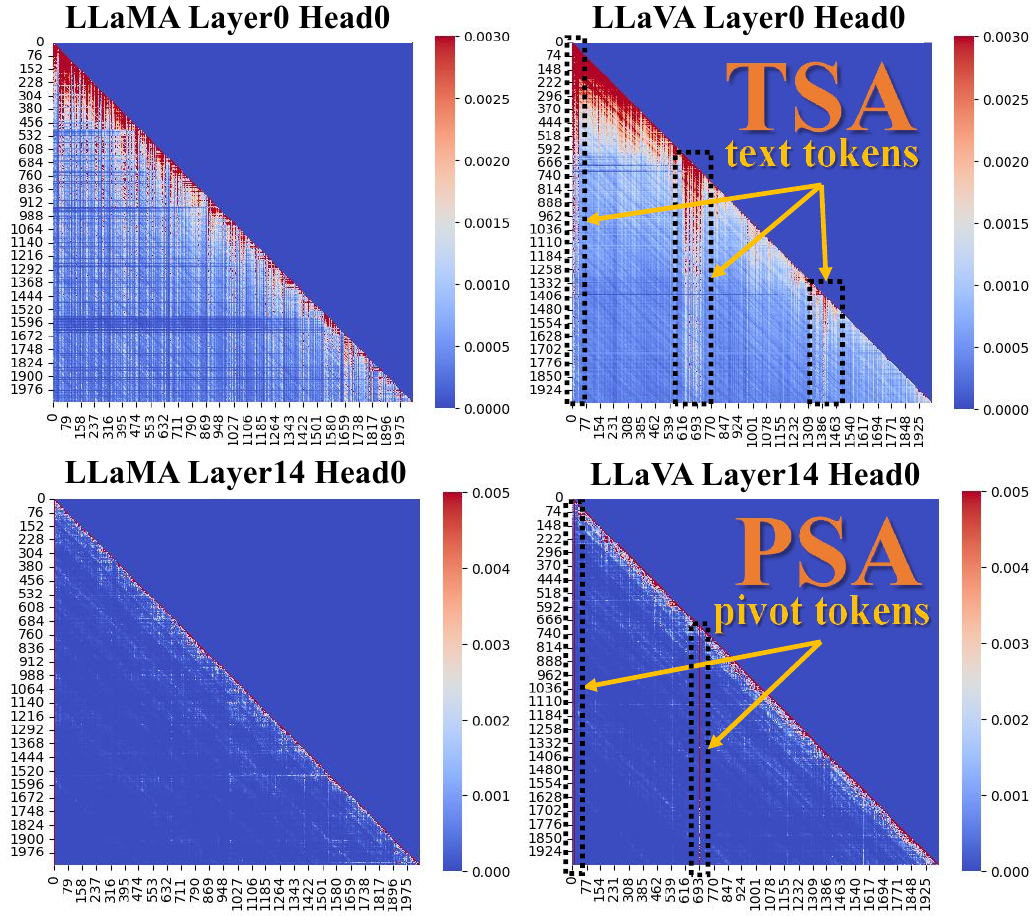}
    \caption{Visualization of representative attention patterns in LLaMA\cite{touvron2023llama} and LLaVA\cite{liu2024visual}.
    In the initial layers, VLM prioritize text tokens, exhibiting \textbf{Text-Salient Attention (TSA)}. 
    In the subsequent layers, TSA diminishes, transitioning to \textbf{Pivot-Token-Salient Attention (PSA)}, where only a few pivot tokens dominate attention.
}
    \label{fig:main image}
\end{figure}
\begin{figure*}[t]
\centering
\includegraphics[width=1\textwidth]{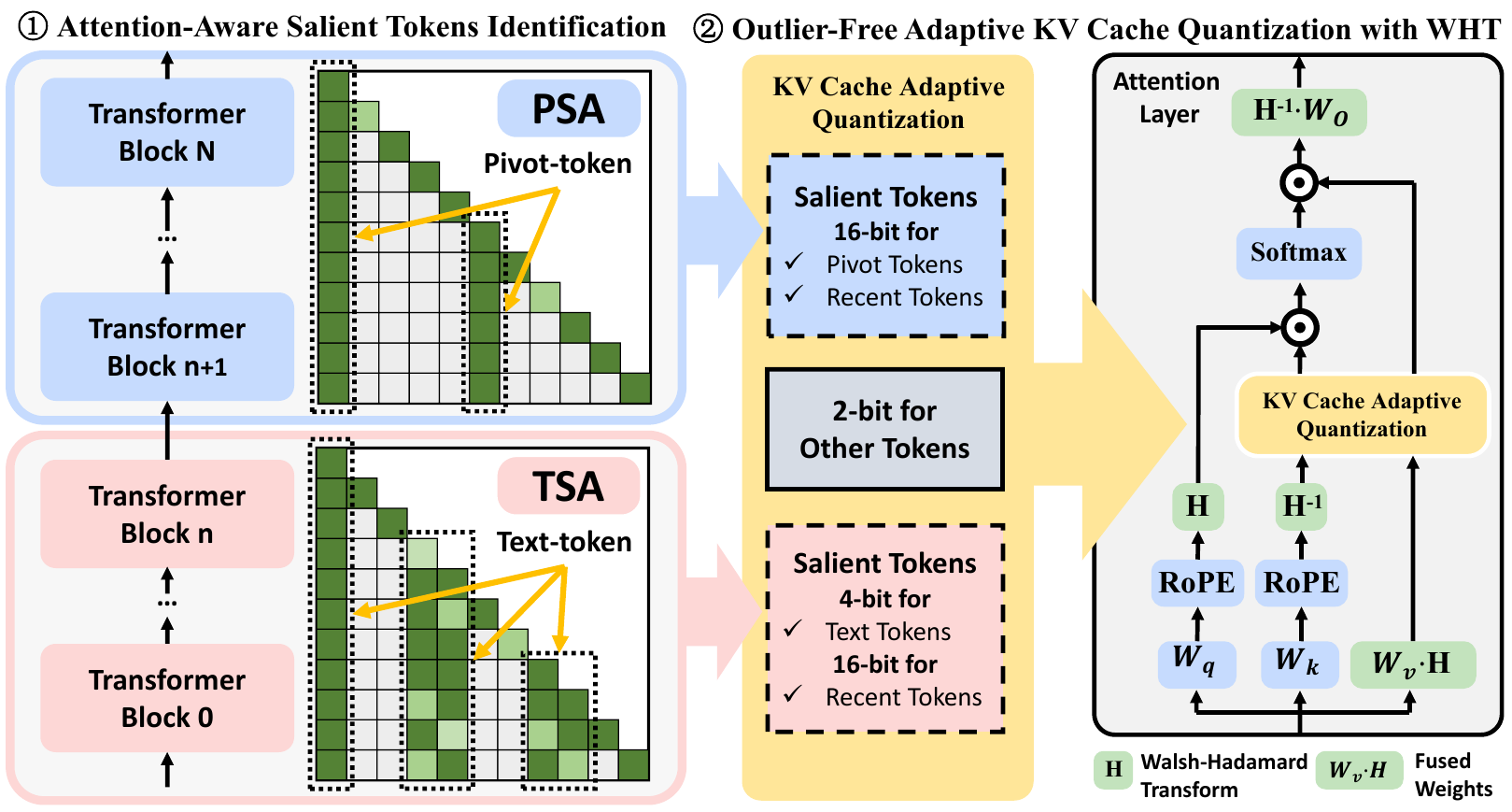} 
      \caption{AKVQ-VL uses an attention-aware technique to identify salient tokens and adaptively quantize the KV cache, with WHT-based equivalent transformations effectively reducing outliers of KV cache.}
    \label{fig:overview}
\end{figure*}
However, as briefly illustrated in Figure \ref{fig:main image}, our observations reveal differences in the patterns of salient tokens within the attention mechanism of VLM compared to LLM.
Directly applying these LLM-oriented methods to VLM overlooks the inherent differences in multimodal KV caches, leading to inefficient compression and suboptimal performance in downstream tasks after quantization.
To the best of our knowledge, no prior research on KV cache quantization has specifically tackled this issue.

In this work, we perform a comprehensive comparative analysis of the attention processes across layers and heads in VLM and LLM, and leverage the attention behaviors of VLM to propose the \textbf{AKVQ-VL} method for optimizing multimodal KV cache quantization.
As shown in Figure \ref{fig:main image}, in the initial layers, VLM prioritize text tokens over vision tokens, a pattern we refer to as \textbf{Text-Salient Attention (TSA)}.
In the subsequent layers, TSA diminishes, transitioning seamlessly to a pattern where attention predominantly concentrates on a small subset of tokens.
Prior study\cite{liu2024intactkv} refers to these few tokens, which receive the majority of attention, as pivot tokens. 
Previous research \cite{xiao2023efficient} also suggests that these tokens typically appear at the beginning of the sequence. 
In contrast, we observe that in VLM, they can also appear at other positions, including within vision tokens.
We refer to this attention pattern as \textbf{Pivot-Token-Salient Attention (PSA)}.
The transition from TSA to PSA suggests that VLM first ``glance'' at the text, prioritizing textual information, and then focus on several pivot tokens.
These findings highlight the inherent differences in KV caches across multimodal tokens and attention layers in VLM, suggesting that traditional LLM-based methods are unable to fully address these challenges.

To address this gap, we propose \textbf{AKVQ-VL}, which first identifies salient tokens based on attention patterns (TSA and PSA), then applies adaptive mixed-precision quantization with the Walsh-Hadamard transform (WHT) to effectively preserve these tokens, while quantize the remaining tokens to lower bit-widths to improve compression efficiency.
AKVQ-VL extends the conventional approach—typically focused on retaining only the initial few tokens during quantization \cite{duanmu2024skvq, liu2024intactkv, hooper2024kvquant}—by leveraging the correlation between massive activations \cite{sun2024massive} and pivot tokens, thereby enabling the effective identification of additional pivot tokens.
Moreover, a key challenge in applying extremely low-bit quantization is the presence of outliers in KV tensors, particularly along the channel dimension of the Keys\cite{liu2024kivi,hooper2024kvquant}. 
To mitigate this issue, we incorporate the WHT into the attention computation, facilitating the construction of an outlier-free KV cache.
Our contributions are summarized as follows:

\textbullet\  \textbf{AKVQ-VL is the first approach specifically designed for KV cache quantization in VLM.} 
Our method leverages the attention patterns of VLM to optimize multimodal KV quantization, achieving nearly lossless 2-bit quantization.

\textbullet\  Through an extensive comparative analysis of attention processes across layers and heads in both VLM and LLM, \textbf{we identify the TSA and PSA patterns in VLM, along with their transitions across layers.}
These insights provide valuable guidance for KV cache compression in VLM. 
Additionally, \textbf{AKVQ-VL constructs an outlier-free KV cache using the WHT, facilitating extremely low-bit quantization.}

\textbullet\  Evaluations across 12 long-context and multimodal tasks on multiple VLMs demonstrates that \textbf{AKVQ-VL maintains or even improves performance with 2-bit quantization}, outperforming previous LLM-oriented methods. 
With the current implementations, AKVQ-VL achieves a 2.13× reduction in peak memory usage, supports up to 3.25× batch sizes, and boosts throughput by 2.46× on LLaVA-v1.5-7B\cite{liu2024visual}. 
\begin{figure*}[t]
\centering
\includegraphics[width=1\textwidth]{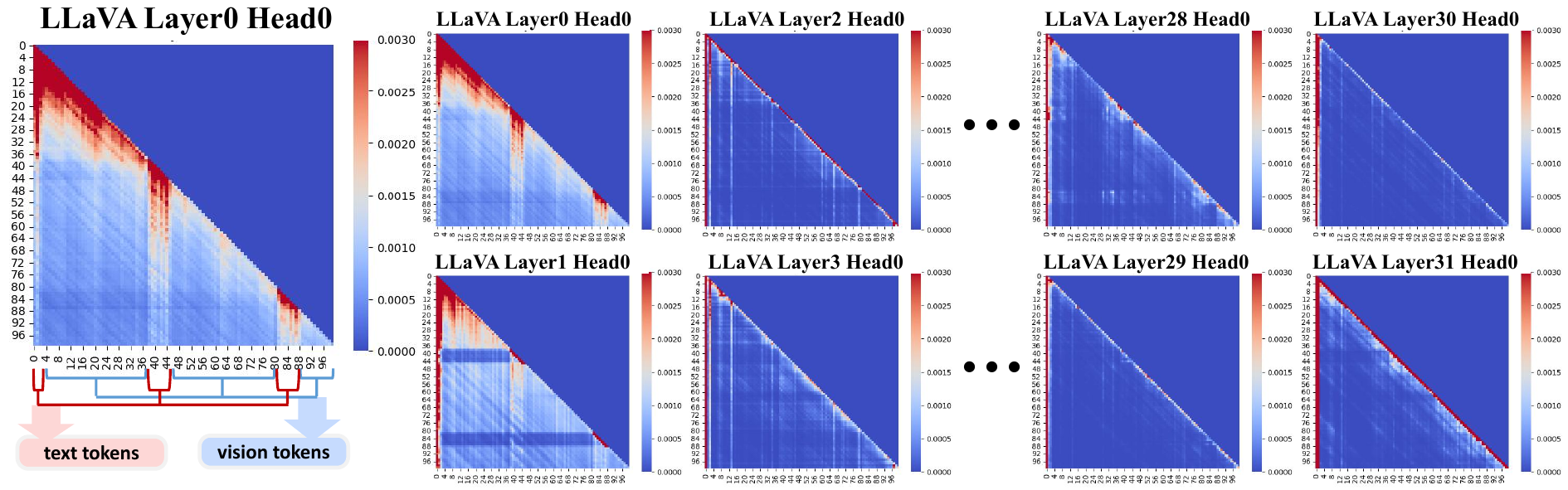} 
      \caption{Visualization of \textbf{average attention scores} in LLaVA\cite{liu2024visual}, averaged across every 8 tokens.
      The first two layers exhibit the TSA pattern, while other layers display the PSA pattern.}
    \label{fig:all attn}
\end{figure*}
\section{Background}
\subsection{VLM Inference with multimodal KV Cache}
The inference process consists of two stages: the prefill phase and the decoding phase:
\subsubsection{\textbf{Prefill Phase}} 
The model processes the token sequence generated from the prompt and generates the initial output token, with each attention layer computing and caching KV pairs. 
Let $\mathbf{X}\in{\mathbb{R}^{l_{prompt}\times d}}$ represent the input embeddings, where $l_{prompt}$ is the length of the input token sequence and $d$ is the model's hidden size. 
In each attention layer, the KV can be derived as follows:
\begin{equation}
    K = \mathbf{X} \cdot W_k, \quad V = \mathbf{X} \cdot W_v,
\end{equation}
where $W_k, W_v\in\mathbb{R}^{d\times d}$ are the weight matrices for the Key and Value calculations, respectively.
\subsubsection{\textbf{Decoding Phase}}
The model takes a single token as input.
Let $\mathbf{t}\in{\mathbb{R}^{1\times d}}$ as the input embedding. 
Each attention layer computes \( t_K \) and \( t_V \) as follows:
\begin{equation}
    t_K = \mathbf{t} \cdot W_k, \quad t_V = \mathbf{t} \cdot W_v.
\end{equation}
Then, \( t_K \) and \( t_V \) are employed to update the KV cache, with the complete KV cache supporting subsequent computations.

Unlike LLM, which only process tokens from a single textual modality, VLM handle both vision tokens from the vision encoder and text tokens from the tokenizer, with the KV cache storing historical information from multimodal inputs.
\section{Methodology}
In Section \ref{3.1}, we analyze the attention process in VLM, identifying distinct patterns that emerge across multiple VLMs and differ from those in LLMs.
We then introduce AKVQ-VL, which integrates two core strategies: attention-aware salient tokens identification and outlier-free adaptive KV cache quantization with WHT, as detailed in Sections \ref{3.2} and \ref{3.3}.
An overview of AKVQ-VL method is shown in Figure \ref{fig:overview}.
\subsection{Distinct Attention Patterns in VLM}
\label{3.1}
We analyze the attention process in VLM using prompts that contain both text and multiple images from MileBench\cite{song2024milebench} with LLaVA-v1.5-7B \cite{liu2024visual} serving as a case study to illustrate our findings. 
Additional VLMs are summarized later.
For comparison, we also analyze attention process of LLaMA2-7B \cite{touvron2023llama}. 
The following summarizes our observations:
\subsubsection{\textbf{Local Attention Pattern}}
\label{local attention pattern}
As illustrated in Figure \ref{fig:main image}, \textbf{both VLM and LLM exhibit a ‘local’ attention pattern, with greater focus on recent tokens.}
Additionally, attention is relatively dispersed across tokens in the initial layers but becomes concentrated on a small set of tokens in the subsequent layers.
This aligns with prior research\cite{xiao2023efficient} on LLM and is anticipated, given that VLM integrate an LLM backbone.
\subsubsection{\textbf{TSA Pattern}}
\label{TSA pattern}
\begin{figure}[t]
    \centering
    \begin{subfigure}[b]{0.48\linewidth}
        \centering
        \includegraphics[width=\linewidth]{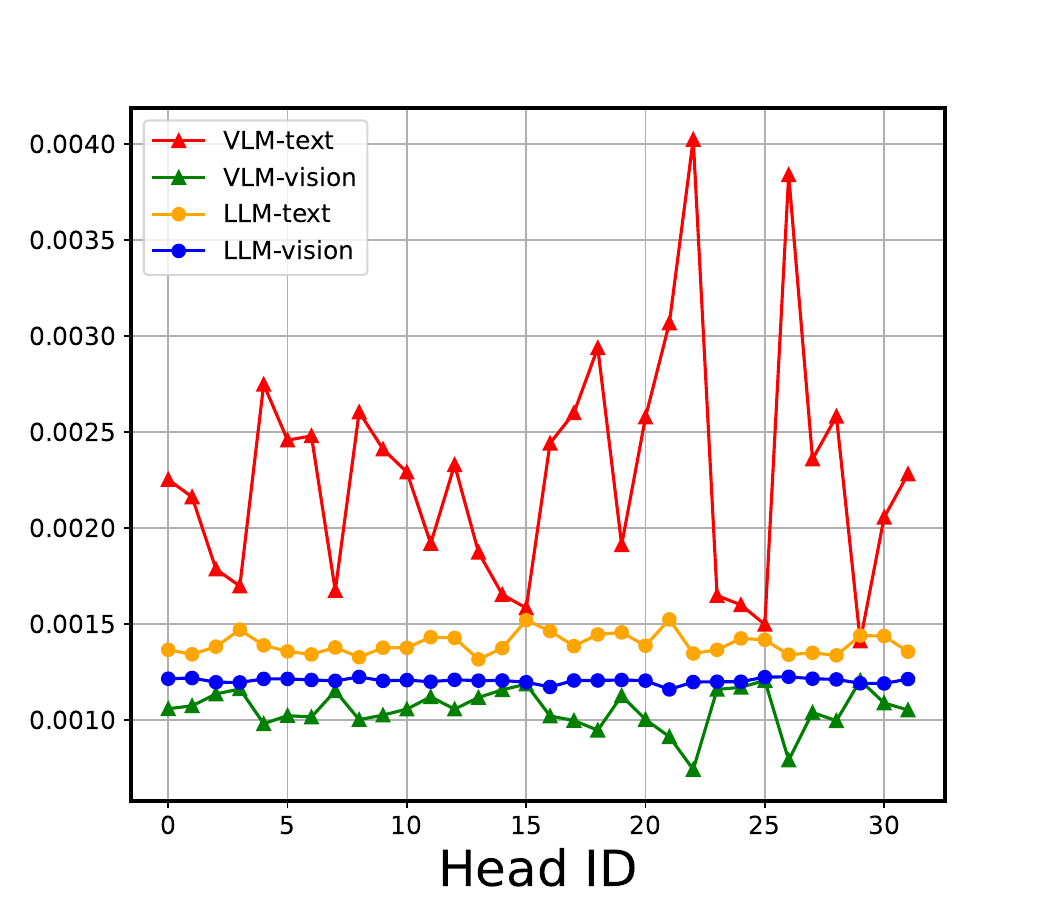}
        \caption{Attention scores of layer 0}
        \label{fig:TSA_a}
    \end{subfigure}
    \hfill
    \begin{subfigure}[b]{0.48\linewidth}
        \centering
        \includegraphics[width=\linewidth]{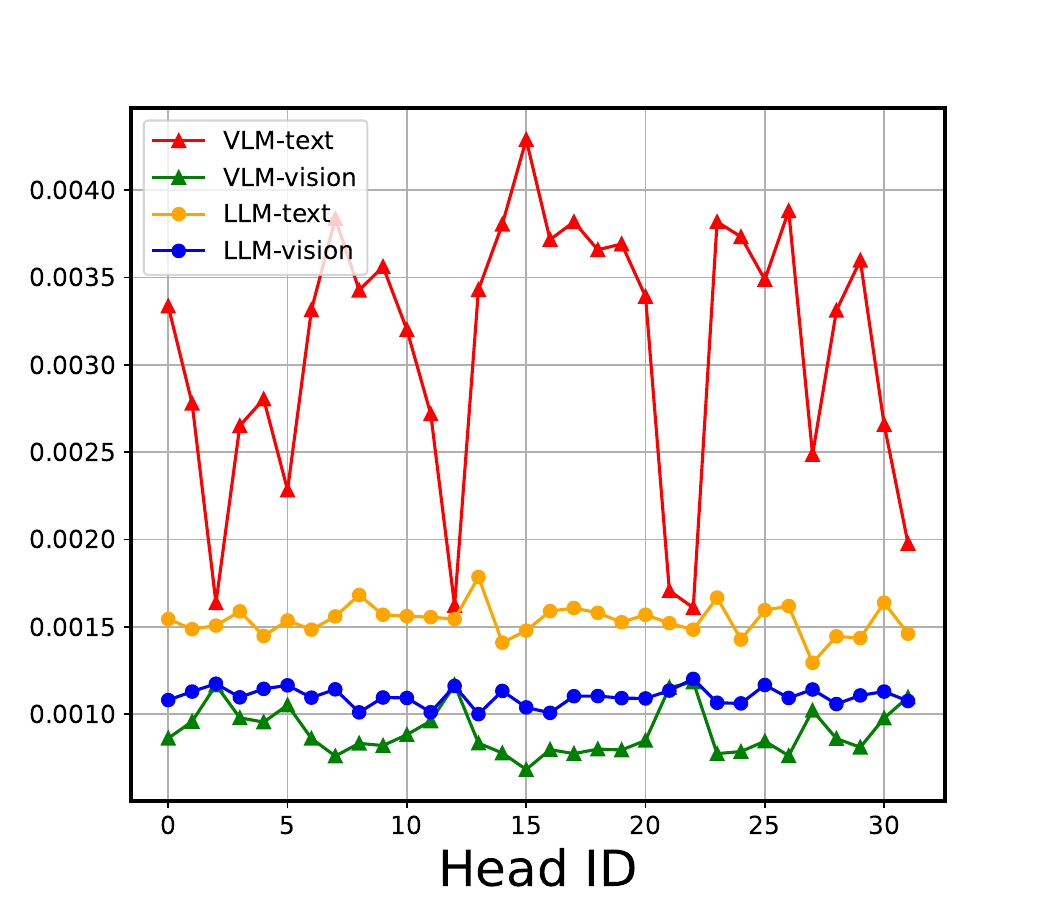}
        \caption{Attention scores of layer 1}
        \label{fig:TSA_b}
    \end{subfigure}
    \caption{Visualization of average attention scores for tokens from different modalities. 
    To mitigate the influence of sink tokens, we exclude the first five tokens. 
    For LLM, we apply the same grouping of indices used in VLM.}
    \label{fig:TSA}
\end{figure}

As shown in Figure \ref{fig:all attn}, in the first two attention layers, \textbf{text tokens dominate attention over vision tokens, illustrating the TSA pattern.}
Despite comprising the majority of the sequence, vision tokens receive comparatively less attention. 
This indicates that the model prioritizes text understanding, using its contextual cues to extract information from the redundant vision tokens.
To validate TSA across attention heads in Layers 0 and 1, we group tokens by modality and compute the average attention scores for each modality.
In the comparison experiment of LLM, we apply the same grouping of indices used in VLM.
From Figure \ref{fig:TSA}, we can conclude that most attention heads in the VLM exhibit varying degrees of TSA.
In contrast, there is no significant difference  between tokens from different groups in LLM.
\subsubsection{\textbf{PSA Pattern}}
\label{PSA pattern}
In subsequent layers, TSA diminishes, and \textbf{pivot tokens begin to dominate the attention, illustrating the PSA pattern.}
As shown in Figure \ref{fig:main image}, in contrast to previous studies\cite{xiao2023efficient,liu2024intactkv}, we observe that in VLM, pivot tokens can appear at positions other than the beginning of the sequence.

In summary, \ref{local attention pattern} demonstrates the presence of ``local'' attention pattern in VLM, similar to that observed in LLM. 
\ref{TSA pattern} highlights the distinct behaviors of text and vision tokens, advocating for separate treatment during the quantization process.
\ref{PSA pattern} reveals patterns in the occurrence of pivot tokens in VLM, highlighting the need to focus on these additional pivot tokens.
Table \ref{tab:table1} summarizes the TSA and PSA patterns observed across various VLMs, including LLaVA-V1.5-13B\cite{liu2024visual}, LLaVA-v1.6-vicuna-7b, LLaVA-v1.6-mistral-7b\cite{liu2023improved} and Qwen2-VL-7B-Instruct\cite{Qwen2VL}.
\begin{table}[t]
\centering
\caption{TSA and PSA patterns across several VLMs.}
\resizebox{0.7\columnwidth}{!}{%
\begin{tabular}{@{}lcc@{}}
\toprule
\multicolumn{1}{c}{\textbf{Model}} & \textbf{TSA Layer} & \textbf{PSA Layer} \\ \midrule
LLaVA-v1.5-7B & 0-1 & 2-31 \\
LLaVA-v1.5-13B & 0-1 & 2-31 \\
LLaVA-v1.6-vicuna-7B & 0-1 & 2-31 \\
LLaVA-v1.6-mistral-7B & None & 0-31 \\
Qwen2-VL-7B & 0-1 & 2-27 \\ \bottomrule
\end{tabular}%
}
\label{tab:table1}
\end{table}
\subsection{Attention-Aware Salient Token Identification}
\label{3.2}
In AKVQ-VL, salient tokens are identified through two key attention patterns: TSA and PSA.
These tokens are then quantized with higher precision, while the remaining tokens are quantized to 2 bits to optimize compression efficiency.
\subsubsection{\textbf{Salient Tokens of TSA}}
\label{3.2.1}
To identify salient tokens, some existing methods \cite{yang2024no,wan2024look} rely on attention scores as a metric, followed by techniques such as KV cache pruning or mixed-precision quantization. 
However, these approaches require access to the attention score during inference, making them less suitable for kernel-based attention acceleration implementations, such as FlashAttention \cite{dao2022flashattention}, which directly outputs the attention results.
Unlike the challenge of efficiently identifying salient tokens in LLM, treating text tokens as salient in VLM is straightforward and naturally aligns with the observed TSA pattern.
Additionally, in line with \ref{local attention pattern}, we adopt established practices \cite{duanmu2024skvq,liu2024kivi} by leveraging the locality of attention to designate recently generated tokens as salient.
\subsubsection{\textbf{Salient Tokens of PSA}}
\label{3.2.2}
Pivot tokens have been shown to be critical for the performance\cite{sun2024massive}.
Due to \ref{PSA pattern}, existing approaches\cite{duanmu2024skvq,hooper2024kvquant,xiao2023efficient} that focus on protecting sink tokens overlook pivot tokens at other positions during VLM inference.
Recent research on massive activations \cite{sun2024massive}—those activations in the residual sums of Transformer block outputs with significantly larger magnitudes than others—suggests that attention is concentrated on these activations.
Specifically, when massive activations occur, the corresponding tokens attract concentrated attention in the subsequent attention layers, forming pivot tokens. 
Therefore, by identifying the token indices of massive activations, additional pivot tokens can be pinpointed. 
Our method first detects massive activations, and then leverages the corresponding token indices to locate the positions of the pivot tokens.
Under the PSA pattern, the identified pivot tokens, along with the recent tokens, are designated as salient tokens.
\subsection{Outlier-Free Adaptive KV Cache Quantization with WHT}
\label{3.3}
After identifying the salient tokens, AKVQ-VL employs adaptive KV cache quantization to preserve the critical KV pairs while efficiently compressing the remaining tokens. 
The integration of WHT effectively mitigates the impact of outliers, facilitating extremely low-bit quantization.
\begin{figure}[t]
    \centering
    \begin{subfigure}[b]{0.45\linewidth}
        \centering
        \includegraphics[width=\linewidth]{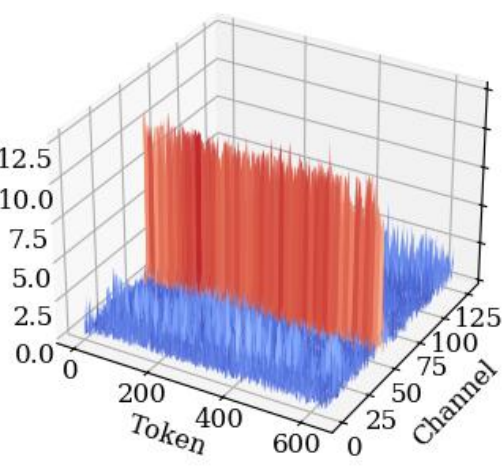}
        \caption{Before WHT}
        \label{fig:outlier_before}
    \end{subfigure}
    \hfill
    \begin{subfigure}[b]{0.45\linewidth}
        \centering
        \includegraphics[width=\linewidth]{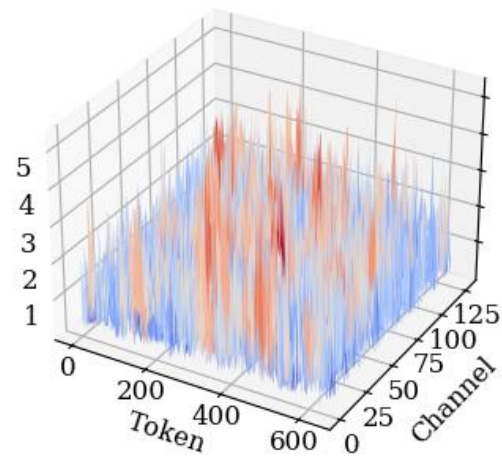}
        \caption{After WHT}
        \label{fig:outlier_after}
    \end{subfigure}
    \caption{The magnitudes of Keys before and after WHT in LLaVA\cite{liu2024visual}, layer 10, head 0.}
    \label{fig:outlier}
\end{figure}
\subsubsection{\textbf{Adaptive Per-Token Dynamic KV Cache Quantization}}
In AKVQ-VL, salient tokens can be classified into two categories. 
The first category comprises pivot and recent tokens, which are limited in number.
The KV associated with these tokens are cached using the original 16-bit precision, as their impact on compression efficiency becomes negligible as the context length increases.
The second category consists of text tokens, which grow in number as the output length expands, and are stored with 4-bit precision.
AKVQ-VL employs per-token dynamic asymmetric quantization with clipping, which can be expressed as follows:
\begin{equation}
Q(X) = clamp\left(\left\lfloor \frac{X}{{scale}} \right\rceil + zero, 0, 2^n - 1 \right),
\end{equation}
\begin{equation}
X' = scale \cdot (Q(X) - zero),
\end{equation}
\begin{equation}
    scale = \frac{clipped\_max(X) - clipped\_min(X)}{2^n - 1} ,
\end{equation}
\begin{equation}
    zero = - \left\lfloor \frac{clipped\_min(X)}{scale} \right\rceil,
\end{equation}
where $\left\lfloor\cdot\right\rceil$ indicates round operation. 
$Q(X)$ and $X'$ denote the quantized and dequantized values of $X$, respectively. 
The $clamp$ operation constrains the values within a specified range. 
$clipped\_max(X)$ and $clipped\_min(X)$ denote the operations that truncate the maximum and minimum values of $X$. 
\begin{table*}[]
\centering
\caption{Evaluations of AKVQ-VL on MileBench\cite{song2024milebench}.}
\label{tab:main result}
\resizebox{1\textwidth}{!}{%
\begin{tabular}{@{}ccccccccccccccc@{}}
\toprule
\multicolumn{1}{c|}{\textbf{Model}} & \multicolumn{1}{c|}{\textbf{Method}} & \multicolumn{1}{l|}{\textbf{Bits}} & \textbf{OE} & \textbf{OI} & \textbf{MA} & \textbf{EN} & \textbf{SC} & \textbf{ST} & \textbf{SU} & \textbf{WQA} & \textbf{TQA} & \textbf{MQA} & \textbf{SQA} & \textbf{DQA} \\ \midrule
\multicolumn{1}{c|}{} & \multicolumn{1}{c|}{FP16} & \multicolumn{1}{c|}{16} & 53.0 & 46.5 & 47.5 & 33.5 & 36.5 & 73.0 & 64.5 & 59.0 & 47.0 & 68.5 & 43.0 & 45.5 \\
\multicolumn{1}{c|}{} & \multicolumn{1}{c|}{RTN (INT4)} & \multicolumn{1}{c|}{4} & 46.0 & 43.5 & 47.5 & 28.5 & 35.5 & 58.5 & 61.0 & 43.0 & 43.5 & 62.5 & 41.5 & 37.5 \\
\multicolumn{1}{c|}{} & \multicolumn{1}{c|}{RTN (INT2)} & \multicolumn{1}{c|}{2} & 15.5 & 12.0 & 12.0 & 13.5 & 8.0 & 7.5 & 9.0 & 7.0 & 16.0 & 13.5 & 9.5 & 5.5 \\ \cmidrule(l){2-15} 
\multicolumn{1}{c|}{} & \multicolumn{1}{c|}{SmoothQuant} & \multicolumn{1}{c|}{2} & 37.5 & 24.0 & 24.5 & 22.0 & 31.5 & 24.5 & 25.0 & 24.0 & 25.5 & 26.5 & 25.5 & 21.0 \\
\multicolumn{1}{c|}{} & \multicolumn{1}{c|}{KIVI} & \multicolumn{1}{c|}{2} & 18.0 & 16.0 & 24.0 & 20.0 & 21.5 & 16.0 & 25.0 & 13.0 & 18.5 & 27.5 & 19.0 & 18.0 \\
\multicolumn{1}{c|}{} & \multicolumn{1}{c|}{SKVQ} & \multicolumn{1}{c|}{2} & 35.0 & 39.0 & 14.0 & 20.0 & 40.5 & 48.5 & 54.5 & 24.5 & 20.5 & 54.5 & 27.0 & 13.5 \\
\multicolumn{1}{c|}{\multirow{-7}{*}{\begin{tabular}[c]{@{}c@{}}LLaVA-v1.5\\ -7B\end{tabular}}} & \multicolumn{1}{c|}{\textbf{AKVQ-VL}} & \multicolumn{1}{c|}{2} & \textbf{54.0} & \textbf{48.0} & \textbf{48.5} & \textbf{36.0} & \textbf{41.0} & \textbf{78.0} & \textbf{66.5} & \textbf{57.0} & \textbf{46.0} & \textbf{68.0} & \textbf{42.5} & \textbf{46.5} \\ \midrule
\multicolumn{1}{c|}{} & \multicolumn{1}{c|}{FP16} & \multicolumn{1}{c|}{16} & 46.0 & 45.0 & 50.0 & 26.5 & 37.0 & 70.5 & 60.5 & 64.0 & 55.0 & 74.5 & 51.0 & 46.0 \\
\multicolumn{1}{c|}{} & \multicolumn{1}{c|}{RTN (INT4)} & \multicolumn{1}{c|}{4} & 49.0 & 42.0 & 47.5 & 28.5 & 35.0 & 65.5 & 57.5 & 64.0 & 51.5 & 71.0 & 45.0 & 47.0 \\
\multicolumn{1}{c|}{} & \multicolumn{1}{c|}{RTN (INT2)} & \multicolumn{1}{c|}{2} & 27.0 & 12.0 & 19.5 & 11.5 & 20.0 & 8.5 & 19.5 & 6.5 & 15.0 & 19.5 & 14.5 & 13.0 \\ \cmidrule(l){2-15} 
\multicolumn{1}{c|}{} & \multicolumn{1}{c|}{SmoothQuant} & \multicolumn{1}{c|}{2} & 36.0 & 26.5 & 23.0 & 21.0 & 31.0 & 25.0 & 31.0 & 24.0 & 27.0 & 26.5 & 25.5 & 24.5 \\
\multicolumn{1}{c|}{} & \multicolumn{1}{c|}{KIVI} & \multicolumn{1}{c|}{2} & 28.5 & 25.5 & 38.0 & 26.5 & 22.0 & 41.0 & 26.5 & 34.0 & 34.0 & 45.5 & 29.5 & 35.5 \\
\multicolumn{1}{c|}{} & \multicolumn{1}{c|}{SKVQ} & \multicolumn{1}{c|}{2} & \textbf{49.0} & 41.5 & 40.0 & 25.5 & \textbf{40.5} & 45.5 & 57.5 & 45.5 & 30.0 & 48.5 & 31.5 & 30.5 \\
\multicolumn{1}{c|}{\multirow{-7}{*}{\begin{tabular}[c]{@{}c@{}}LLaVA-v1.5\\ -13B\end{tabular}}} & \multicolumn{1}{c|}{\textbf{AKVQ-VL}} & \multicolumn{1}{c|}{2} & 43.5 & \textbf{44.0} & \textbf{50.5} & \textbf{26.5} & 35.5 & \textbf{69.5} & \textbf{61.5} & \textbf{64.0} & \textbf{51.0} & \textbf{75.0} & \textbf{48.5} & \textbf{46.5} \\ \midrule
\multicolumn{1}{c|}{} & \multicolumn{1}{c|}{FP16} & \multicolumn{1}{c|}{16} & 29.0 & 18.0 & 18.5 & 28.0 & 20.0 & 63.0 & 34.0 & 50.5 & 41.5 & 38.5 & 42.0 & 41.5 \\
\multicolumn{1}{c|}{} & \multicolumn{1}{c|}{RTN (INT4)} & \multicolumn{1}{c|}{4} & 33.5 & 8.5 & 30.0 & 18.5 & 17.0 & 61.5 & 39.5 & 27.0 & 33.0 & 22.5 & 27.5 & 30.5 \\
\multicolumn{1}{c|}{} & \multicolumn{1}{c|}{RTN (INT2)} & \multicolumn{1}{c|}{2} & 8.0 & 4.0 & 3.0 & 4.5 & 2.5 & 4.0 & 4.5 & 6.5 & 10.5 & 3.5 & 6.5 & 5.0 \\ \cmidrule(l){2-15} 
\multicolumn{1}{c|}{} & \multicolumn{1}{c|}{SmoothQuant} & \multicolumn{1}{c|}{2} & 35.0 & 13.5 & 23.5 & 21.5 & 17.5 & 26.5 & 28.5 & 25.5 & 24.5 & 28.5 & 24.5 & 24.5 \\
\multicolumn{1}{c|}{} & \multicolumn{1}{c|}{KIVI} & \multicolumn{1}{c|}{2} & 17.5 & 15.5 & \textbf{27.0} & 17.5 & 14.5 & 35.0 & 21.0 & 18.5 & 15.5 & 21.0 & 12.5 & 14.5 \\
\multicolumn{1}{c|}{} & \multicolumn{1}{c|}{SKVQ} & \multicolumn{1}{c|}{2} & 34.0 & \textbf{26.5} & 23.5 & 11.0 & \textbf{32.5} & 32.0 & 19.0 & 31.0 & 11.5 & 33.0 & 23.0 & 23.0 \\
\multicolumn{1}{c|}{\multirow{-7}{*}{\begin{tabular}[c]{@{}c@{}}LLaVA-v1.6\\ -vicuna-7B\end{tabular}}} & \multicolumn{1}{c|}{\textbf{AKVQ-VL}} & \multicolumn{1}{c|}{2} & \textbf{35.5} & 14.5 & \textbf{27.0} & \textbf{32.5} & 20.0 & \textbf{63.5} & \textbf{35.5} & \textbf{48.0} & \textbf{45.0} & \textbf{42.5} & \textbf{43.5} & \textbf{38.5} \\ \midrule
\multicolumn{1}{c|}{} & \multicolumn{1}{c|}{FP16} & \multicolumn{1}{c|}{16} & 44.5 & 46.0 & 44.0 & 34.5 & 38.0 & 71.0 & 74.5 & 55.5 & 49.5 & 66.0 & 44.0 & 38.0 \\
\multicolumn{1}{c|}{} & \multicolumn{1}{c|}{RTN (INT4)} & \multicolumn{1}{c|}{4} & 35.5 & 24.0 & 26.0 & 28.0 & 20.0 & 63.0 & 34.0 & 50.5 & 41.5 & 38.5 & 42.0 & 41.5 \\
\multicolumn{1}{c|}{} & \multicolumn{1}{c|}{RTN (INT2)} & \multicolumn{1}{c|}{2} & 29.0 & 18.0 & 18.5 & 20.0 & 18.0 & 11.0 & 22.5 & 13.5 & 18.0 & 25.5 & 17.5 & 22.5 \\ \cmidrule(l){2-15} 
\multicolumn{1}{c|}{} & \multicolumn{1}{c|}{SmoothQuant} & \multicolumn{1}{c|}{2} & 38.5 & 26.0 & 26.0 & 24.0 & 33.0 & 45.0 & 61.0 & 29.0 & 45.5 & 27.0 & 34.0 & 33.5 \\
\multicolumn{1}{c|}{} & \multicolumn{1}{c|}{KIVI} & \multicolumn{1}{c|}{2} & 45.5 & 42.5 & \textbf{47.5} & 24.0 & 36.0 & 53.0 & 63.5 & 47.0 & 49.5 & 62.5 & 39.0 & 37.5 \\
\multicolumn{1}{c|}{} & \multicolumn{1}{c|}{SKVQ} & \multicolumn{1}{c|}{2} & \textbf{49.0} & 36.5 & 39.5 & 27.5 & 34.5 & 56.0 & 65.0 & 46.5 & 47.5 & 52.5 & 37.0 & 33.5 \\
\multicolumn{1}{c|}{\multirow{-7}{*}{\begin{tabular}[c]{@{}c@{}}LLaVA-v1.6\\ -mistral-7B\end{tabular}}} & \multicolumn{1}{c|}{\textbf{AKVQ-VL}} & \multicolumn{1}{c|}{2} & 46.0 & \textbf{43.5} & 43.5 & \textbf{30.0} & \textbf{37.0} & \textbf{68.5} & \textbf{70.5} & \textbf{54.5} & \textbf{51.5} & \textbf{63.5} & \textbf{46.0} & \textbf{42.0} \\ \bottomrule
\end{tabular}%
}
\end{table*}
\subsubsection{\textbf{WHT-Based Outlier Reduction}}
As shown in Figure \ref{fig:outlier_before}, the Key tensor of VLM exhibit significant outliers along the channel dimension. 
These outliers lead to substantial quantization errors when applying extremely low-bit quantization.
Inspired by recent studies \cite{ashkboos2024quarot,liu2024spinquant} that utilize the WHT to mitigate outliers in LLM, we incorporate WHT-based equivalent transformations into AKVQ-VL. 
This strategy effectively reduces outliers without disrupting the original computation, enabling the construction of an outlier-free KV cache.

\begin{figure}[t]
    \centering
    \includegraphics[width=0.7\linewidth]{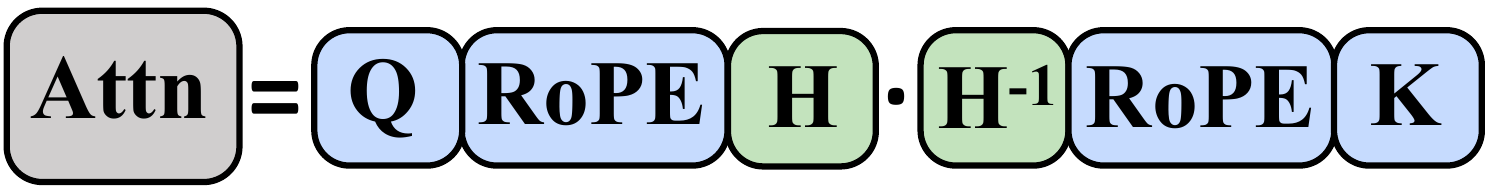}
        \caption{Equivalent transformations for the Key. 
        \( H \) denotes the Walsh-Hadamard matrix and \( Q/K \) represents the Query/Key.
}
        \label{fig:QK}
\end{figure}
\begin{figure}[t]
    \centering
\includegraphics[width=0.7\linewidth]{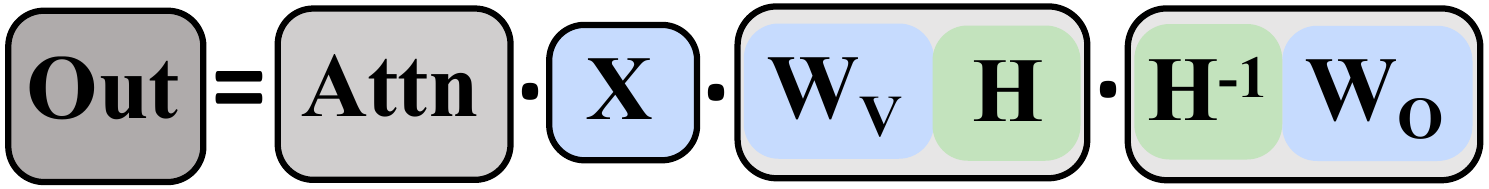}
        \caption{Equivalent transformations for the Value. 
        \( H \) denotes the Walsh-Hadamard matrix, \( X \) represents the input matrix, and \( W_V / W_O \) denotes the weight matrices.
}
        \label{fig:VO}
\end{figure}
Walsh-Hadamard matrix is a specific type of orthogonal matrix characterized by entries proportional to $\{ +1, -1 \}$, and is generated recursively as follows, the subscript denoting the dimension of matrix, where \( k \in \mathbb{Z}^+ \):
\begin{equation}
    H_1 = \begin{bmatrix} 1 \end{bmatrix}, \quad
    H_{2^k} = \frac{1}{\sqrt{2}} \begin{bmatrix}
    H_{2^{(k-1)}} & H_{2^{(k-1)}} \\
    H_{2^{(k-1)}} & -H_{2^{(k-1)}}
    \end{bmatrix}.
\end{equation}
The scaling factor \( \frac{1}{\sqrt{2}} \) ensures normalization. 

The overview of the equivalent transformations in AKVQ-VL is shown in Figure \ref{fig:overview}.
To perform WHT on the Key, we apply the equivalent transformations on-the-fly to both the Query and Key after the Rotary Position Encoding (RoPE) \cite{su2024roformer}, as shown in Figure \ref{fig:QK}.
Notably, the online computational overhead of WHT can be reduced using the Fast Walsh-Hadamard Transform (FWHT) algorithm \cite{fino1976unified}.
Since RoPE does not apply to the Value, the equivalent transformations for the Value can be precomputed by integrating Walsh-Hadamard matrix into \( W_V \) and \( W_o \) offline, as shown in Figure \ref{fig:VO}, thus reducing the online computational overhead.
As illustrated in Figure \ref{fig:outlier_after}, applying the WHT to the Key effectively reduces the outliers.
\begin{figure*}[t]
    \centering
    \begin{subfigure}[b]{0.24\linewidth}
        \centering \includegraphics[width=\linewidth]{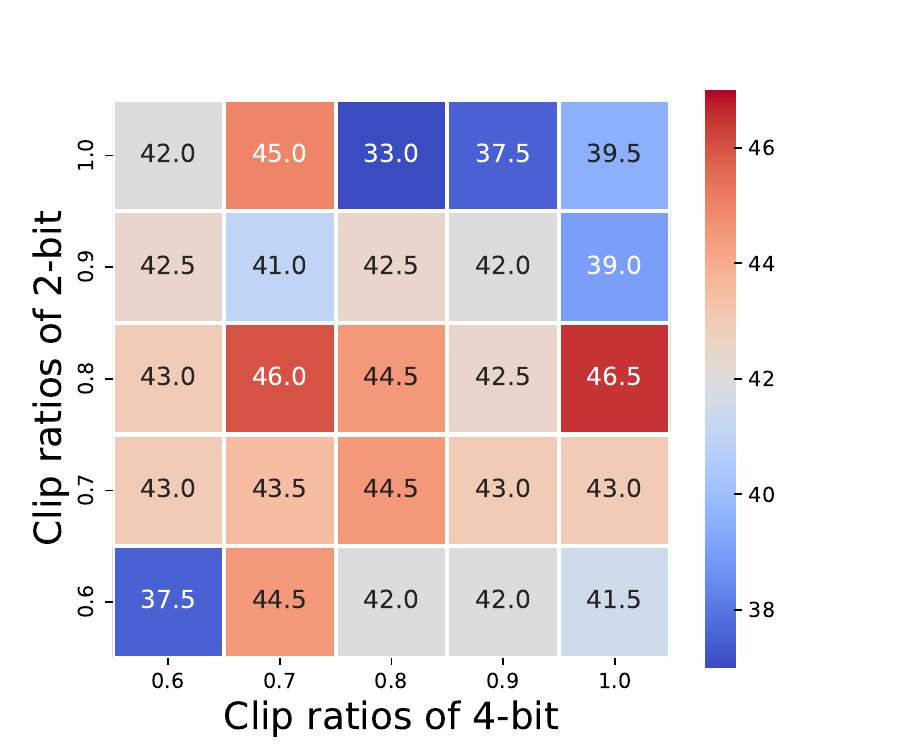}
        \caption{Clip ratios of quantization}
        \label{fig:clip1}

    \end{subfigure}
    \begin{subfigure}[b]{0.24\linewidth}
        \centering      \includegraphics[width=\linewidth]{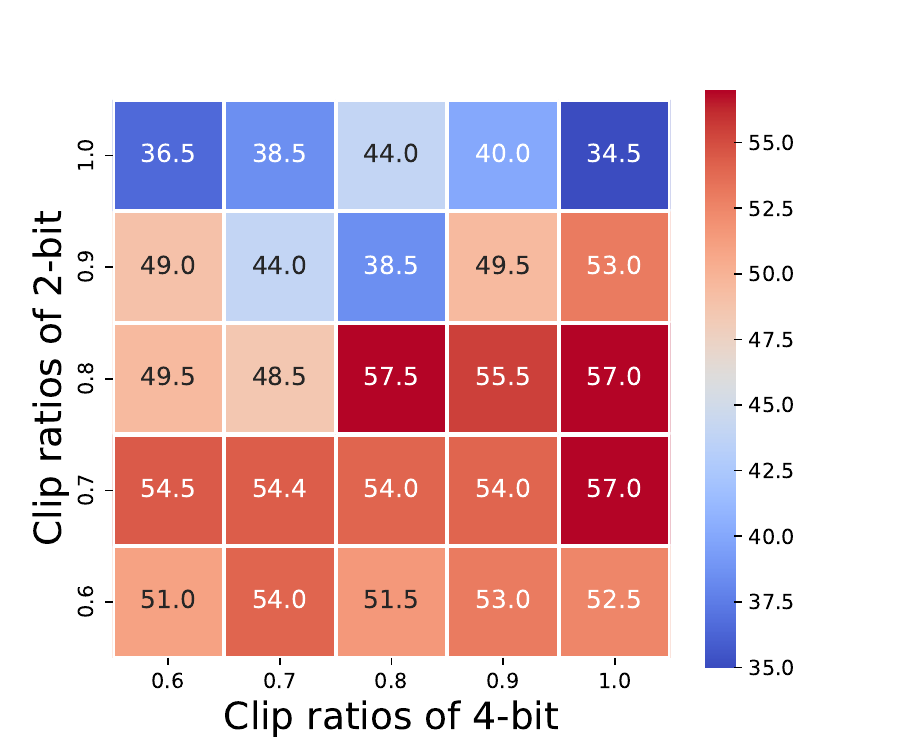}
                
        \caption{Clip ratios of quantization}
        \label{fig:clip2}

    \end{subfigure}
    \begin{subfigure}[b]{0.22\linewidth}
        \centering
    \includegraphics[width=\linewidth]{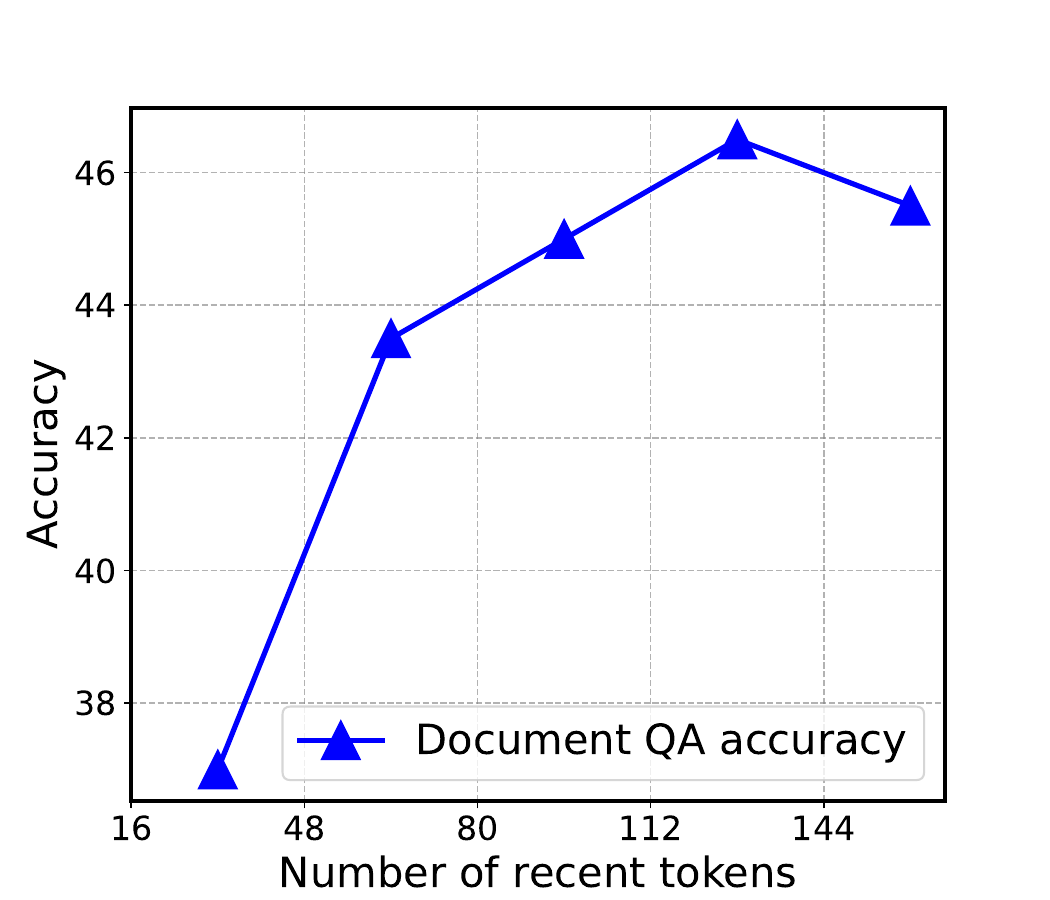}
        \caption{Recent token numbers}
        \label{fig:clip_modality2}
    \end{subfigure}
    \begin{subfigure}[b]{0.22\linewidth}
        \centering
        \includegraphics[width=\linewidth]{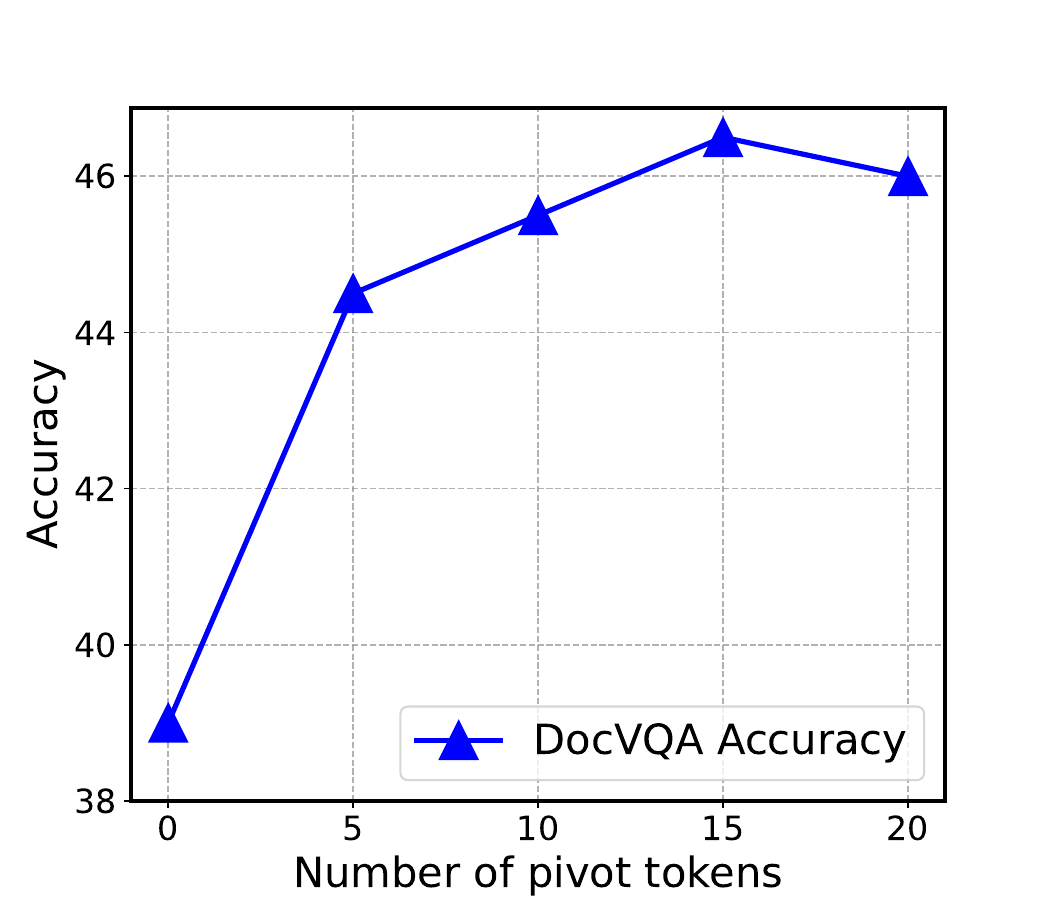}
        \caption{Pivot token numbers}
        \label{fig:clip_modality3}

    \end{subfigure}
    \caption{Visualization of ablation study results.
    }
    \label{fig:clip}
\end{figure*}
\section{Experiments}
\subsection{Experiment Settings}
\subsubsection{\textbf{Models}} We evaluate AKVQ-VL on several VLMs, including LLaVA-v1.5-7B/13B\cite{liu2024visual} and LLaVA-v1.6-vicuna-7B\cite{liu2023improved}, which are based on Multi-Head Attention (MHA), as well as LLaVA-v1.6-mistral-7B\cite{liu2023improved}, which utilizes Grouped Query Attention (GQA).
\subsubsection{\textbf{Tasks and Metrics}}
We evaluate AKVQ-VL using MileBench\cite{song2024milebench}, a comprehensive benchmark for assessing multimodal LLMs on both multi-image and long-context tasks, which aligns with our testing requirements for KV cache compression.
We select 12 tasks from MileBench in a balanced way, including Object Existence (OE), Object Interaction (OI), Moving Attribute (MA), Egocentric Navigation (EN), State Change (SC), Scene Transition (ST), Space Understanding (SU), Webpage QA (WQA), Textbook QA (TQA), multimodal QA (MQA), Slide VQA (SQA), and Document QA (DQA), all using accuracy as the evaluation metric.
\subsubsection{\textbf{Baselines}}
In addition to comparing AKVQ-VL with the uncompressed FP16 and round-to-nearest (RTN) INT4/INT2 quantization KV cache, we further evaluate its advantages in multimodal KV cache quantization by benchmarking it against three state-of-the-art LLM quantization methods: SmoothQuant\cite{xiao2023SmoothQuant}, KIVI\cite{liu2024kivi}, and SKVQ\cite{duanmu2024skvq}. 
SmoothQuant reduces the difficulty of quantization by scaling the activation across channel dimension. 
In our experiments, we apply SmoothQuant to scale the Keys and Values, with the parameter \(\alpha\) set to 0.5.
KIVI accumulates a certain number of FP16 KV caches during decoding phase, then applies per-channel quantization to Keys and per-token quantization to Values.
SKVQ employs a sliding window to store recent KV cache in FP16 and uses a channel reordering technique to reduce quantization difficulty. 
For all methods, the quantization group size is set to 128.
In our experiments, the residual length of KIVI and the window size in SKVQ is all set to 128.
\subsection{Main Results}
As shown in Table \ref{tab:main result}, across 12 multimodal tasks evaluated on multiple VLMs, our method achieves the highest accuracy in most cases, surpassing existing LLM-based approaches and demonstrating superior robustness. 
Notably, despite most of the KV cache in our method being quantized to 2 bits, AKVQ-VL consistently preserves accuracy in downstream tasks and outperforms RTN (INT4) in most cases, even outperforms the FP16 baseline in many cases.
\subsection{Ablation Studies}
\begin{table}[t]
\centering
\caption{Ablation study of the proposed components}
\resizebox{0.96\columnwidth}{!}{%
\begin{tabular}{@{}lr@{}}
\toprule
\textbf{Method} & \textbf{Scene Transition (Accuracy)} \\ \midrule
FP16 & 73 \\ \midrule
RTN (INT2) & 7.5 (65.5 $\downarrow$) \\
+ WHT &  20.5 (13.0 $\uparrow$)\\
+ Adaptive Quantization Based on TSA &  41 (20.5 $\uparrow$)\\
+ Adaptive Quantization Based on PSA & 78 (37.0 $\uparrow$) \\ \bottomrule
\end{tabular}%
}
\label{table:Ablation Studies}
\end{table}
\subsubsection{\textbf{Clip Ratios of Quantization}}
We conduct ablation experiments on clip ratios for both 4-bit and 2-bit quantization, focusing on the Document QA and Webpage QA tasks. 
As shown in Figure \ref{fig:clip1} and \ref{fig:clip2}, the experiments demonstrate that a clip ratio between 0.7 and 0.8 yields optimal performance for 2-bit quantization, while 4-bit quantization exhibits less sensitivity to the clip ratio.
Therefore, we choose stable configurations of 0.8 for 2-bit quantization and 1.0 for 4-bit quantization.
\subsubsection{\textbf{Numbers of Recent Tokens and Pivot Tokens}}
We conduct ablation experiments on the Document QA task to examine the impact of the number of recent tokens and pivot tokens in PSA. 
As shown in Figures \ref{fig:clip_modality2} and \ref{fig:clip_modality3}, to achieve the optimal balance between accuracy improvement and compression efficiency, we set the number of recent tokens to 128 and the number of pivot tokens in PSA to 15.
\subsubsection{\textbf{Ablation Study of the Proposed Components}}
Starting with a naive RTN (INT2) quantization, we progressively integrate the proposed components and evaluate accuracy on the Scene Transition task. 
As shown in Table \ref{table:Ablation Studies}, the innovations in AKVQ-VL effectively mitigate performance degradation.
\begin{table}[t]
\centering
\caption{Efficiency analysis}
\label{tab:Efficiency Analysis}
\resizebox{0.8\columnwidth}{!}{%
\begin{tabular}{@{}cccc@{}}
\toprule
\textbf{Method} & \textbf{Batch Size} & \textbf{\begin{tabular}[c]{@{}c@{}}Memory Usage\\ (GB)\end{tabular}} & \textbf{\begin{tabular}[c]{@{}c@{}}Throughput\\ (tokens/s)\end{tabular}} \\ \midrule
FP16 & 40 & 63.08 & 1028.38 \\
AKVQ-VL & 130 & 62.57 & 2526.37 \\ \bottomrule
\end{tabular}%
}
\end{table}
\subsection{Efficiency Analysis}
\begin{figure}[t]
    \centering
\includegraphics[width=0.6\linewidth]{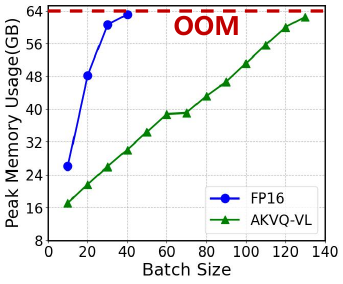}
    \caption{Peak memory usage test. 
}
    \label{fig:runtime}
\end{figure}

In this section, we evaluate the efficiency of AKVQ-VL. 
To reduce the overhead of dynamic quantization, we implement the quantization and dequantization using Triton. 
Additionally, we implement FWHT using an optimized CUDA kernel following QuaRot\cite{ashkboos2024quarot}. 
We evaluate LLaVA-v1.5-7B\cite{liu2024visual} on a 4-NVIDIA V100 (16GB) setup, using an input length of approximately 500 tokens.
The batch size is increased progressively until an out-of-memory (OOM) error occurs, and we report the peak memory usage and throughput. 
As shown in Table \ref{tab:Efficiency Analysis} and Figure \ref{fig:runtime}, AKVQ-VL achieves a 2.13× reduction in peak memory usage, supports up to 3.25× larger batch sizes, and boosts throughput by 2.46×. 
It is worth noting that throughput can be further enhanced through techniques such as kernel fusion.
\section{Conclusion}
In this paper, through a comprehensive comparative analysis of the attention process in VLM, we propose AKVQ-VL, the first approach specifically designed for KV cache quantization in VLM. 
AKVQ-VL adaptively compresses the multimodal KV cache and applies WHT to reduce the impact of outliers, achieving nearly lossless 2-bit quantization.
Our experiments show that AKVQ-VL reduces memory usage and enhances throughput  while maintaining strong performance on downstream tasks. 
Future work will focus on further optimizing AKVQ-VL for even greater efficiency.

\end{document}